Muhammad Burhan Hafez*, Cornelius Weber, Matthias Kerzel, and Stefan Wermter

# Deep intrinsically motivated continuous actor-critic for efficient robotic visuomotor skill learning

**Abstract:** In this paper, we present a new intrinsically motivated actor-critic algorithm for learning continuous motor skills directly from raw visual input. Our neural architecture is composed of a critic and an actor network. Both networks receive the hidden representation of a deep convolutional autoencoder which is trained to reconstruct the visual input, while the centre-most hidden representation is also optimized to estimate the state value. Separately, an ensemble of predictive world models generates, based on its learning progress, an intrinsic reward signal which is combined with the extrinsic reward to guide the exploration of the actor-critic learner. Our approach is more data-efficient and inherently more stable than the existing actor-critic methods for continuous control from pixel data. We evaluate our algorithm for the task of learning robotic reaching and grasping skills on a realistic physics simulator and on a humanoid robot. The results show that the control policies learned with our approach can achieve better performance than the compared state-of-the-art and baseline algorithms in both dense-reward and challenging sparse-reward settings.

## 1 Introduction

An autonomous agent learning control skills from trial and error in an unknown environment with zero prior knowledge is faced with the challenging task of correctly mapping the often high-dimensional sensory observations to motor actions. Reinforcement Learning (RL) allows for learning such a mapping by finding action policies that maximize future rewards from the environment. In recent years, deep RL has achieved great success in solving several control problems, utilizing deep neural networks as powerful nonlinear function approximators [1]. However, deep RL suffers from poor sample efficiency as it requires large amounts of training data and the agent needs to actively collect it online, rendering it generally impractical for real-world robotic learning. In this article, we make two novel contributions: First, a deep autoencoder architecture is proposed that aids the learning of deep RL parameters through online joint optimization of supervised and unsupervised objectives. Second, we derive an efficient exploration strategy using the learning progress of an ensemble of predictive models. Combined, this leads to the Deep Intrinsically motivated Continuous Actor-Critic (Deep ICAC) algorithm which optimizes model-free policy learning and provides model-based intrinsic feedback towards accelerating real robot learning. We evaluate our Deep ICAC algorithm for learning-to-reach and learning-to-grasp tasks on simulated and real robots.

### 1.1 The problem of data efficiency in deep RL

To improve sample efficiency in deep RL, different approaches have recently been proposed. Schaul *et al.* pointed out that for most deep RL methods, transitions are randomly drawn from a replay buffer of recent transitions whenever a learning update for the network weights is performed. Instead of this inefficient sampling, they proposed a Prioritized Experience Replay, where each transition in the buffer is assigned a sampling probability proportional to its temporal-difference error [2]. High priority is thus given to samples of large errors, and thus of a high potential for updating the network weights, making experience replay more efficient. In a different approach, an agent learns an estimate of the expectation over the future state representations from a given state and action, called

***Corresponding Author: Muhammad Burhan Hafez:** Department of Informatics, University of Hamburg, Germany;
E-mail: hafez@informatik.uni-hamburg.de
**Cornelius Weber, Matthias Kerzel, Stefan Wermter:** Department of Informatics, University of Hamburg, Germany;
E-mail: {weber, kerzel, wermter}@informatik.uni-hamburg.de

successor representation (SR) [3]. This allows for replacing the state-action value function, which estimates the expected future reward, with a function estimating only the immediate reward using the SR, and thereby eliminating the need for the slow propagation of state-action values among visited states.

Another study suggests that in order to enhance the exploration efficiency, the uncertainty about the state-action values needs to be propagated back when updating the value estimates instead of only the scalar mean values as usually done in previous works [4]. For that, a probability distribution is defined over returns from a state-action pair and approximated with a deep neural network, which, along with its parametric uncertainty, is propagated using the Bellman equation. The policy is then improved with Thompson sampling based on both the parametric uncertainty of the network and the distributional uncertainty of the return. Integrating model-based and model-free learning by training a model-free RL agent on real as well as model-generated, imagined trajectories has also shown to improve data efficiency [5].

Other studies demonstrate that unsupervised learning from self-generated reward leads to efficient exploration. In [6], an exploration incentive based on the prediction error of a learned model of the environmental dynamics is used as internal feedback and added to the observed reward. The incentive encourages visiting novel states and is applied as an alternative to the count-based exploration bonuses, which are impractical in large domains [6]. Similarly, Jaderberg et al. show that maximizing auxiliary rewards representing percept*ual cha*nges on the sensory as well as the learned feature levels while learning the target task makes the training faster. To achieve this, they train agents to learn separate policies that maximize the perceived changes in image pixel values and in neuron activations of each layer of the value and policy networks, optimizing the combined loss of the auxiliary and the base agents [7].

While the above approaches offer a variety of techniques in which data efficiency in deep RL can be improved, they are limited to discrete action domains unsuitable in a realistic robotic setting. To address continuous control with deep RL, a few attempts have been made over the last two years. For example, Deep Deterministic Policy Gradient (DDPG), a state-of-the-art deep RL algorithm, has been successfully applied to continuous control tasks [8]. It learns an action-value function in an off-policy manner from trajectories generated by a stochastic behavior policy and updates the deterministic target policy by gradient ascent on the value function. More recently, Kalweit and Boedecker showed that DDPG's high sample complexity can be reduced by performing a learning update on not only real but also imaginary transition samples generated by a learned dynamics model [9]. As opposed to [5], imaginary samples are not used each time an update is performed but based on an uncertainty measure derived by bootstrapping the critic's neural network. Similar to DDPG, Asynchronous Advantage Actor-Critic (A3C) is an actor-critic policy gradient algorithm, but learns a stochastic instead of deterministic target policy. A3C asynchronously updates the deep network parameters of multiple agents in parallel and has shown to improve learning efficiency [10].

## 1.2 The role of state representation learning in RL

In realistic high-dimensional sensory space, it is particularly helpful for an RL agent to learn task-relevant features that make learning the desired control behavior easier. Learning good state representation in RL thus has received wide attention in recent years. For instance, using autoencoders to learn compact low-dimensional state representations unsupervised for RL has been proposed [11–13]. However, a common limitation to these methods is that they require a separate pre-training phase to adjust the autoencoder weights prior to learning the policy for the target task. Therefore, they learn features that do not necessarily distinguish rewarding states. Unsupervised learning of temporally coherent features has shown to provide invariant representations that also improve the learning speed of RL agents in different control tasks [14, 15]. These works use Slow Feature Analysis (SFA) as an unsupervised method for learning invariances from temporal input sequences. While these approaches learn low-dimensional feature abstraction that is more biologically plausible and noise robust than the abstraction learned with autoencoders, they are susceptible to learning task-irrelevant slow features and perform expensive eigenvalue decomposition which is also done before starting to optimize the action policy.

## 1.3 RL and sparse feedback

Besides the need to learn in continuous action spaces, truly autonomous agents need to learn how to act when extrinsic rewards are only sparsely available. In order to allow RL agents to efficiently and meaningfully explore in a sparse-reward world, intrinsically motivated RL methods have been proposed providing the agent a number of intrinsic drives, with artificial curiosity being the most common. Different functions have been defined to design an intrinsic reward for the agent, including Bayesian surprise [16], information gain [17], empowerment [18], prediction error of a learned forward model [6, 19, 20], predictive learning progress [21, 22], and policy value change [23]. In deep RL, Stadie et al. and Jaderberg et al. also use intrinsic feedback to aid the learning of a target task instead of relying exclusiv*ely on* sparse extrinsic rewards [6, 7]. However, their proposed measures for computing the intrinsic reward are mainly based on the perceptual novelty of the observed states, that is highly sensitive to noise commonly found in real-world systems.

In another approach, intrinsic rewards are used in hierarchical deep RL to encourage a low-level controller to reach an intrinsic goal state chosen by a high-level controller that learns a policy over intrinsic goals to optimize some extrinsic reward from the environment [24]. Despite being limited to discrete actions, the proposed algorithm was shown to significantly outperform the DQN algorithm of Mnih *et al.* [1] in sparse reward tasks with a complex goal structure. More recently, intrinsic feedback has been applied in self-play between two copies of the same agent where the first periodically sets a goal for the second to achieve and is intrinsically rewarded proportionally to the time taken by the second to complete the task [25]. The second is rewarded inversely proportional to the time taken to complete the first one's chosen task. Moving from self-play to target task learning, a learning speed-up was shown when the second copy was deployed to solve the target task.

In this paper, we use an intrinsic reward based on the learning progress of a growing ensemble of predictive models, that is less sensitive to noise and accords with the surprise-enhanced learning [26], where violation-of-expectation, seen as a prediction error, enhances children's learning, and the Goldilocks effect principle in infant cognition that attributes optimal learning to stimuli of an intermediate difficulty [27]. This is evident in how infants seek increasingly complex learning samples by selectively shifting their interactions with the world from well-explored regions to others where they expect to learn new effects of motor activity acquiring information of events that are neither too predictable nor too surprising.

# 2 Deep intrinsically motivated continuous actor-critic (deep ICAC)

Our approach to learning goal-directed continuous control policies involves two interacting parts: (1) training the actor and critic networks with experience replay based on our deep variant of the Continuous Actor-Critic Learning Automaton (CACLA) algorithm [28]; and (2) incorporating predictive model-ensemble intrinsic reward for directed exploration. We first give the necessary background on continuous actor-critic RL including the CACLA algorithm in Section 2.1 and then describe our proposed learning architecture for Deep CACLA which we use as the RL controller in Section 2.2. We present our Deep ICAC algorithm in Section 2.3.

## 2.1 Continuous actor-critic RL

We consider a standard finite-horizon MDP where an agent takes an action $a_t$ from its action space $A$ in a state $s_t$ from its state space $S$ each timestep $t$ and observes a new state $s_{t+1}$ and reward $r_t$. This transition is described by a state transition model mapping from a state-action pair to a distribution over $S$. When acting in its environment the agent executes a policy $\pi : S \to P(A)$ mapping from a state to a distribution over $A$. A return from a state is defined as the total discounted reward $R_t = \sum_{i=t}^{T} \gamma^{i-t} r(s_i, a_i)$ with a discount factor $\gamma \in [0, 1]$. The value function is the expected return from state $s$, $V^\pi(s) = E[R_t \mid s_t = s, \pi]$when following a policy $\pi$. The goal of the RL agent is to find an optimal policy maximizing the expected return:

$$\pi^* = \arg max_\pi E_{s \sim S_0} \left[V^\pi(s)\right] \tag{1}$$

where $S_0 \subseteq S$ is a set of initial states.

To solve Eq. 1, a true transition model is necessary. However, in complex domains, learning a good transition model is often computationally expensive and severely limits the policy by the accuracy of the learned model and thus model-free RL methods can be used in such cases. In model-free, value-based RL, an action-value function is defined as the expected return from taking action $a$ in state $s$ and following policy $\pi$ hereafter, $Q^\pi(s, a) = E[R_t \mid s_t = s, a_t = a, \pi]$, and can be computed recursively using the Bellman equation: $Q^\pi(s, a) = E_{s_{t+1} \sim M}\left[r(s_t, a_t) + \gamma E_{a_{t+1} \sim \pi}\left[Q^\pi(s_{t+1}, a_{t+1})\right] \mid s = s_t, a = a_t\right]$. In continuous state spaces, a function approximator is used to learn parameters $\theta^Q$ that minimize the

loss:

$$L\left(\theta^{Q}\right)=\left(y_t^{Q}-Q^{\pi}\left(s_t,a_t\mid\theta^{Q}\right)\right)^2 \tag{2}$$

where $y_t^Q = r(s_t, a_t) + \gamma \max_a Q^{\pi}\left(s_{t+1}, a \mid \theta^Q\right)$. The value-function approximator is updated by gradient descent on the loss in Eq. 2.

Actor-critic methods are a class of RL algorithms that learn a value function and a policy simultaneously and have shown promising results with the advances in deep learning. They are particularly suitable for continuous action spaces. In a standard actor-critic RL, the actor suggests an action $a_t$ in state $s_t$ and the critic evaluates the action utility through the observed reward $r_t$ and next state $s_{t+1}$, and using this evaluation, the actor refines its future suggestions. Deep Deterministic Policy Gradient (DDPG) [8] and CACLA [28] are two popular continuous actor-critic methods.

### 2.1.1 DDPG

DDPG is a model-free actor-critic algorithm that learns a deterministic target policy $\mu$ from transitions generated by an arbitrary stochastic policy. The policy $\mu$ directly maps states to actions and represents the current approximation of the optimal policy. The critic estimates the action-value function. Function approximators $\mu\left(s \mid \theta^{\mu}\right)$ and $Q\left(s, a \mid \theta^{Q}\right)$ are used to estimate the actor and critic with parameters $\theta^{\mu}$ and $\theta^{Q}$ respectively. The critic is updated by the Bellman equation using slowly changing target value and policy networks found to stabilize learning in the Deep Q-Networks algorithm [1]. A random minibatch of $n$ transitions of the form $(s_i, a_i, s_{i+1}, r_i)$ is first drawn and corresponding targets $y_i$ are computed using the target value $Q^{'}\left(s, a \mid \theta^{Q^{'}}\right)$ and policy $\mu^{'}\left(s \mid \theta^{\mu^{'}}\right)$ networks, $y_i = r_i + \gamma Q^{'}\left(s_{i+1}, \mu^{'}\left(s_{i+1} \mid \theta^{\mu^{'}}\right) \mid \theta^{Q^{'}}\right)$. A minibatch stochastic gradient descent (SGD) step is then performed on the loss function $L = \frac{1}{n}\sum_i \left(y_i - Q\left(s_i, a_i \mid \theta^Q\right)\right)^2$ to update the parameters $\theta^Q$. The actor is updated in the direction of the sampled policy gradient:

$$\nabla_{\theta^{\mu}} J = \frac{1}{n}\sum_i \nabla_a Q\left(s, a \mid \theta^{Q}\right)|_{s=s_i,\, a=\mu(s_i)} \nabla_{\theta^{\mu}}\mu\left(s \mid \theta^{\mu}\right)|_{s=s_i} \tag{3}$$

where $J$ is a performance objective representing the expected return of the target policy and $n$ is the minibatch size. The parameters of the target networks are also moved slowly toward their corresponding parameters of the policy and value networks, $\theta^{Q^{'}} \leftarrow \tau\,\theta^{Q} + (1-\tau)\,\theta^{Q^{'}}$ *and* $\theta^{\mu^{'}} \leftarrow \tau\,\theta^{\mu} + (1-\tau)\,\theta^{\mu^{'}}$, with $\tau \ll 1$.

### 2.1.2 CACLA

Like DDPG, CACLA is a model-free actor-critic algorithm. The critic learns a parameterized approximation of the state-value function by applying a Temporal-Difference (TD) learning update, $V_{t+1}\left(s_t \mid \theta^V\right) = V_t\left(s_t \mid \theta^V\right) + \alpha_t\delta_t$, where $\delta_t = r_t + \gamma V_t\left(s_{t+1} \mid \theta^V\right) - V_t\left(s_t \mid \theta^V\right)$ is the TD-error and $\alpha_t \in [0, 1]$ is the learning rate. The parameters $\theta^V$ are updated by an SGD on the loss $\frac{1}{2}(\delta_t)^2$, which moves the value estimate closer to $r_t + \gamma V_t\left(s_{t+1} \mid \theta^V\right)$. The actor is represented by a function approximator $Ac\left(s \mid \theta^{AC}\right)$. In contrast to DDPG, the actor here is only updated when the TD-error is positive. The reason is that when an explored action $a_t$ results in an increase in the critic's estimate of the state $s_t$ value, then this action is believed to lead to potentially higher future rewards and thus the target policy is updated in the direction of that action. The actor's parameters $\theta^{AC}$ are adjusted by performing a conditional SGD update on the loss $\frac{1}{2}\left(a_t - AC\left(s_t|\theta^{AC}\right)\right)^2$ as follows:

$$If\ \delta_t > 0:$$

$$\theta^{AC} \leftarrow \theta^{AC} + \alpha\left(a_t - AC\left(s_t|\theta^{AC}\right)\right)\nabla_{\theta^{AC}}AC\left(s_t|\theta^{AC}\right) \tag{4}$$

No update is performed when the value estimate is not actually improving (*i.e.* $\delta_t \le 0$) because otherwise that would update toward an action that might not be better than the currently known best action. This update rule is a major difference to the policy gradient algorithms that do not consider the distance to the promising action but the size of the value improvement. By using the sign of the TD-error rather than its size when updating the actor's policy, CACLA is more invariant to scaling of rewards.

## 2.2 Deep CACLA

For high-dimensional state spaces, the actor and critic require good representations capable of identifying states that lead to high future rewards in order to learn a good value function which makes learning the desired policy easier. To support this, we propose an architecture that includes learning a low-dimensional feature representation $\phi_{s_t}$ using a Convolutional Autoencoder (CAE). The CAE is jointly trained with the critic's neural network $V\left(.|\omega,\ \theta^V\right)$ that outputs an estimate of the expected state value using the features $\phi_{s_t}$ computed by the convolutional encoder $f$ with parameters $\omega$. The decoder $g$ with parameters $\tilde{\omega}$ decodes $\phi_{s_t}$ into the state space. The actor is represented by a separate neural network that takes in $\phi_{s_t}$ and outputs a current estimate of the best action. The architecture is shown in Figure 1.

The CAE learns the encoder parameters $\omega$ and decoder parameters $\tilde{\omega}$ that minimize the L2 reconstruction loss between the input image $s_t$ and the reconstructed image $\hat{s}_t$:

$$L_t^c\ (\omega, \tilde{\omega}) = \left(g\left(\phi_{s_t}|\ \tilde{\omega}\right) - s_t\right)^2 \tag{5}$$

The loss function for the critic's parameters is given by:

$$L_t^v\left(\omega, \theta^V\right) = E\left[\left(y_t - V\left(\phi_{s_t}|\omega,\ \theta^V\right)\right)^2\right] \tag{6}$$

where $y_t$ is the target value calculated using the target value network $V^{'}\left(.|\omega^{'},\ \theta^{V^{'}}\right)$ and the reward $r_t$ observed when acting at the state $s_t$ and equals $r_t + \gamma V^{'}\left(\phi_{s_{t+1}}|\omega^{'},\ \theta^{V^{'}}\right)$. A target network with slowly updated parameters is often used in deep value-based RL to provide more stationary targets, as mentioned in Section 2.1.1. For learning the actor's parameters, when the TD-error is positive (see Eq. 4), the following loss function is used:

$$L_t^a\left(\omega, \theta^{AC}\right) = \left(a_t - AC\left(\phi_{s_t}|\ \theta^{AC}\right)\right)^2 \tag{7}$$

where $a_t$ is the exploratory action taken at $s_t$.

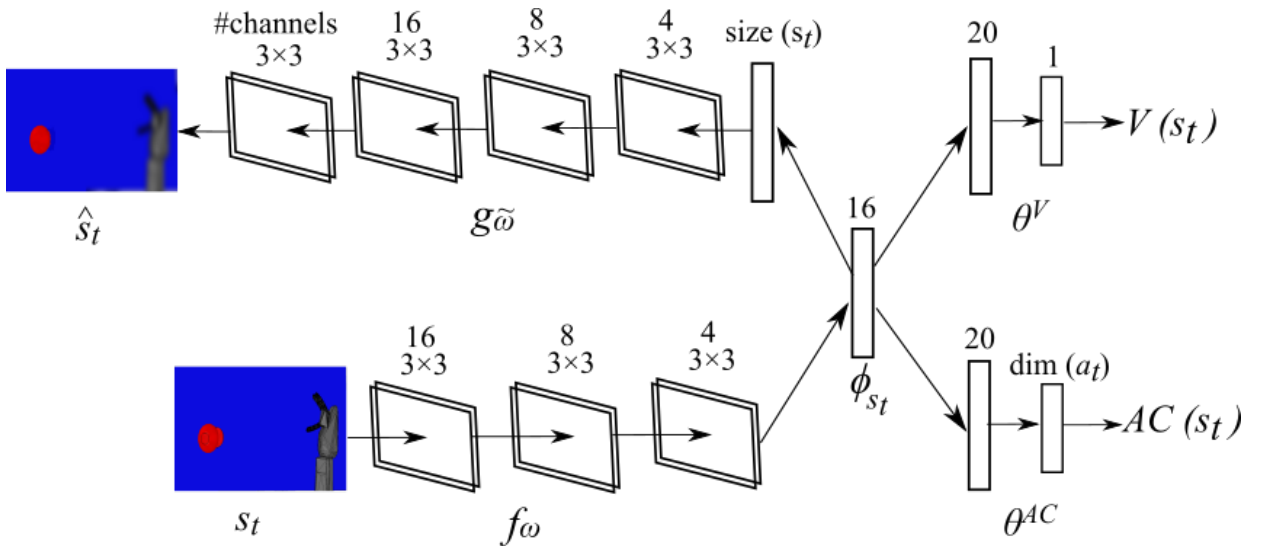

**Figure 1:** Learning Architecture: The architecture consists of (1) a convolutional encoder branch $f_\omega$ that takes in a raw image $s_t$ and extracts a feature vector $\phi_{s_t}$, (2) a convolutional decoder branch $g_\omega$ that produces a reconstruction $s_t$ of the input, (3) a value branch $V$ with parameters $\theta^V$ that estimates the expected value using the features $\phi_{s_t}$, and (4) a policy branch $AC$ with parameters $\theta^{AC}$ that outputs a current estimation of the best action.

The proposed deep model is trained online with minibatch SGD to find values for the parameters $(\omega, \tilde{\omega}, \theta^V, \theta^{AC})$ that minimize the combined loss:

$$L_t\left(\omega, \tilde{\omega}, \theta^V, \theta^{AC}\right) = L_t^c\ (\omega, \tilde{\omega}) + L_t^v\left(\omega, \theta^V\right) + L_t^a\left(\omega, \theta^{AC}\right) \tag{8}$$

To optimize Eq. 8 with respect to the learning parameters $\{\omega, \tilde{\omega}, \theta^V, \theta^{AC}\}$, we perform an iterative update of the parameters $\{\omega, \tilde{\omega}, \theta^V\}$ to minimize $L_t^c\ (\omega, \tilde{\omega}) + L_t^v\left(\omega, \theta^V\right)$ and the parameters $\theta^{AC}$ to minimize $L_t^a\left(\omega, \theta^{AC}\right)$ after fixing $\{\omega, \tilde{\omega}, \theta^V\}$. This update ensures that the parameters $\omega$ are not affected by backpropagation of $L_t^a$ gradients since we want the actor to take the learned $\phi_{s_t}$ as its input and not $s_t$. We also move the target value network parameters slowly toward the learned network parameters as follows: $\theta^{V^{'}} \leftarrow \tau\,\theta^V + (1-\tau)\,\theta^{V^{'}}$, $\omega^{'} \leftarrow \tau\omega + (1-\tau)\,\omega^{'}$ with $\tau \ll 1$. The minibatch is drawn from a replay buffer of size $10^5$ to perform a minibatch SGD on the combined loss (Eq. 8) at each update iteration.

The feature representation $\phi_{s_t}$ in the proposed architecture is learned to be a good state representative and value predictor by sharing the parameters $\omega$ between the encoder and the critic, which is highly desirable for efficient learning of good action policies. The learning system is shown in Figure 2.

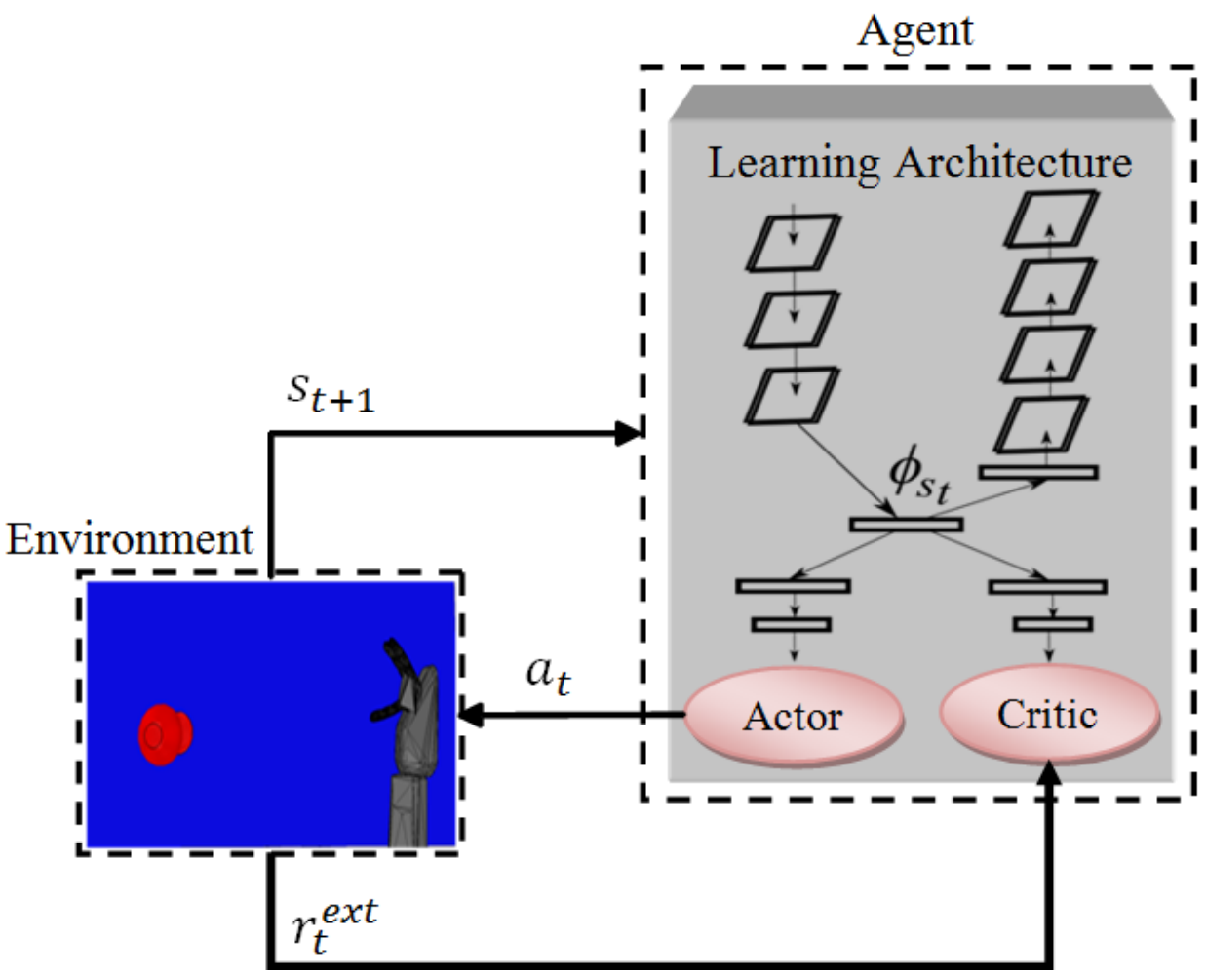

**Figure 2:** Deep CACLA learning system: The convolutional encoder of the agent's learning architecture computes a feature representation $\phi_{s_t}$ in state $s_t$. The agent then takes an action $a_t$ chosen by the actor network based on $\phi_{s_t}$ and the environment returns a new state $s_{t+1}$ and reward $r_t^{ext}$ used to update the critic's estimated utility of $a_t$. Finally, the actor is updated towards $a_t$ if it is found to improve the critic's estimated utility (i.e., TD-error > 0).

## 2.3 Deep ICAC

While learning a policy in continuous action space, balancing exploration and exploitation becomes significantly challenging. Simply following a randomized exploration approach would be highly inefficient in such a space. Instead, to achieve a more efficient and directed exploration, we propose an intrinsic reward function based on the learning progress of an ensemble of local predictive models of the world dynamics.

### 2.3.1 Predictive model-ensemble intrinsic reward

Two key principles to our approach to intrinsic reward design are predictive learning progress and self-organization of sensory space. This is inspired by how infants continually organize their interaction with the world as they learn about its dynamics, shifting their focus from explored to unexplored regions driven by curiosity.

In our approach, the state space is incrementally partitioned into local regions using a growing self-organizing map model $M$. Since the RL agent explores its sensory space along continuous trajectories, the Instantaneous Topological Map (ITM) model [29] is used as our self-organizing map model $M$. It is designed for strongly correlated stimuli and is simpler and grows faster than other growing self-organizing models. The ITM is defined by a set of nodes $i$, each with a weight vector $w_i$, and a set of edges connecting each node $i$ to its neighbors $N(i)$. The ITM starts with two connected nodes, and whenever a new state $s$ is observed (here the state is represented by its feature vector $\phi_s$), the following adaptation steps are performed:

1. Matching: Find the nearest node $n$ and the second nearest node $n'$ to the observed state $\phi_s$: $n \leftarrow argmin_i \left\| \phi_s - w_i \right\|$, $n' \leftarrow argmin_{j,j \neq n} \left\| \phi_s - w_j \right\|$.
2. Edge adaptation: Create an edge between $n$ and $n'$ if they are not connected. Check, for all nodes $m$ in $N(n)$, whether $n'$ lies inside the Thales sphere through $m$ and $n$ (*i.e.* $(w_n - w_{n'}) \cdot (w_m - w_{n'}) < 0$). If this is true, remove the edge between $n$ and $m$, and then, if $m$ has no remaining edges, remove $m$.
3. Node adaptation: If $\phi_s$ lies outside the Thales sphere through $n$ and $n'$, *i.e.* $(w_n - \phi_s) \cdot (w_{n'} - \phi_s) > 0$, and if the distance between $n$ and $\phi_s$ is greater than a given threshold $e_{max}$, add a new node $v$ with a weight vector $w_v = \phi_s$ and an edge with $n$.

Each region of the state space (node in $M$) is assigned a local predictive model $p$ trained to predict the next state, given the current state and action. Then, the change between two consecutive average prediction errors of a predictor associated with the best-matching node $n$ in $M$ for the current state is computed:

$$LP_t = \left| \left\langle e_t^{prd} \right\rangle - \left\langle e_{t-T}^{prd} \right\rangle \right| \tag{9}$$

where $T$ is a time window and $\left\langle e_t^{prd} \right\rangle$ is the average prediction error computed over the $\mu$ recent predictions,

$$\left\langle e_t^{prd} \right\rangle = \frac{1}{\mu} \sum_{i=1}^{\mu} e_i^{prd} \,|_{e_i^{prd} = \left\| P\left(\phi_{s_i}, a_i\right) - \phi_{s_{i+1}} \right\|}$$

This change represents the learning progress $LP_t$ the agent has made or expects to make and is combined with the perception error $e_t^{per}$ which is the distance between the state encoding $\phi_{s_t}$ and the weight vector of node $n$ to give an intrinsic reward signal:

$$r_t^{int} = LP_t + e_t^{per} \tag{10}$$

This self-generated reward encourages the agent to try actions that are expected to maximize its learning progress and to lead to perceptually novel states. In this way, the agent is not solely attracted to states with large prediction error (*i.e.* high novelty) which could attract it to noisy states that retain a large prediction error. This intrinsic reward also provides information on which regions of the state space the agent is less certain about its action outcomes and thus exploration is required. Being locally defined, the intrinsic reward facilitates moving from well-explored to less explored regions of the world, which is also suitable for locally structured domains where actions are defined only on parts of the environment.

To use the derived intrinsic reward in our proposed actor-critic model, we gradually anneal it to account for the fact that with more interactions the agent becomes less uncertain about its world dynamics. We combine it with the extrinsic reward as follows:

$$r_t = r_t^{ext} + \frac{r_t^{int}}{1 + D \cdot t} \tag{11}$$

where $D > 0$ is a decay constant. Figure 3 shows the overall learning system, demonstrating the interaction among the different components of our approach at one timestep. The learning algorithm is detailed in Algorithm 1.

# 3 Experiments and results

We evaluate our approach on robotic learning-to-reach and learning-to-grasp tasks. In all the experiments, we compare the proposed Deep ICAC to our Deep CACLA baseline and state-of-the-art DDPG. We consider two environmental conditions for each task: dense-reward and sparse-reward settings. The hyperparameter settings used in all the experiments are discussed in Section 3.1. Results are then presented in Section 3.2.

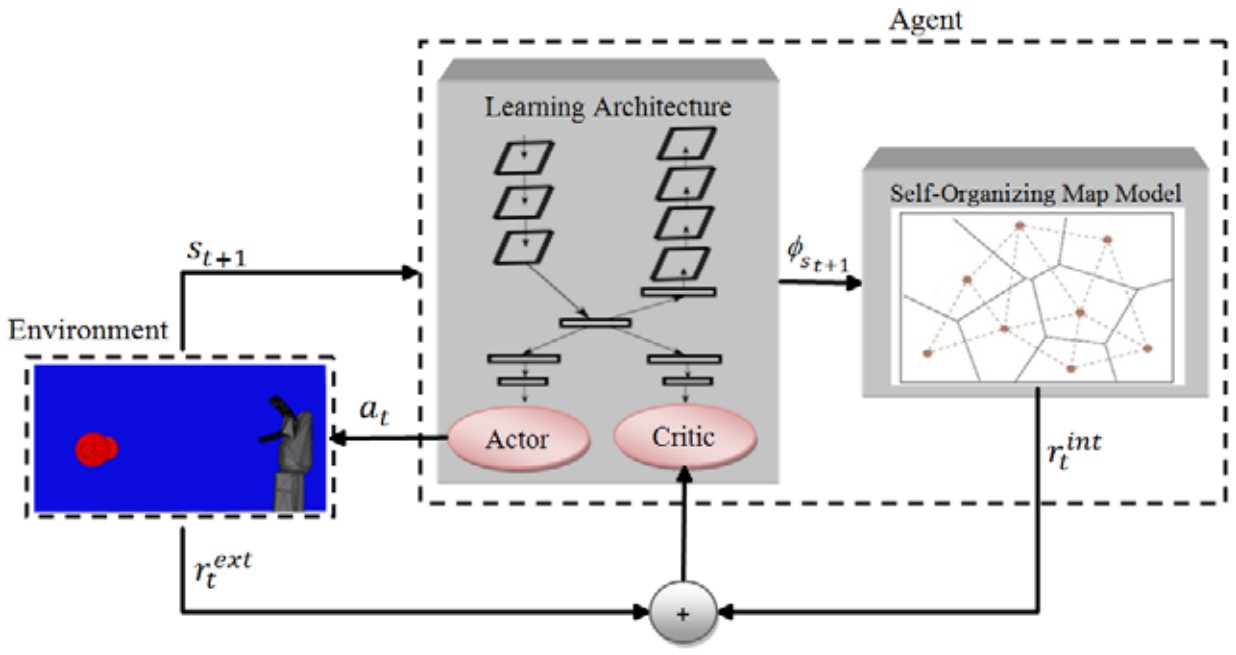

**Figure 3:** Deep ICAC learning system: The agent takes an action $a_t$ chosen by the actor in state $s_t$ and the environment returns a new state $s_{t+1}$ and reward $r_t^{ext}$. The convolutional encoder of the agent's learning architecture then computes a feature representation $\phi_{s_{t+1}}$ which the self-organizing map model M uses to adapt its topology and, along with the action $a_t$, update the learning progress of the predictor corresponding to the M's best-matching node for $\phi_{s_t}$. The updated learning progress is used to derive an intrinsic reward $r_t^{int}$ that is combined with the extrinsic reward $r_t^{ext}$, if any, and fed to the critic to update its estimate of the utility of $a_t$. Finally, the actor is updated towards $a_t$ if it is found to improve the critic's estimated utility.

## 3.1 Parameters

We employ a convolutional autoencoder that includes 7 zero-padded convolutional layers with ReLU activations, 2 dense layers with ReLU activations, and no pooling layers, as shown in the encoder and decoder branches of Figure 1. The figure also shows the number and size of the filters used in each layer. All convolutional layers have the same filter size (3×3) applied with stride 1 to maintain the size of the input image. The critic network consists of the encoder layers followed by a dense layer with 20 ReLU neurons and a dense output layer of a single linear neuron. The fourth layer of the encoder is a dense layer with 16 neurons whose output is used as a low-dimensional feature vector $\phi$ and fed to the actor network. The actor network is a 2-layer fully-connected MLP of 20 tanh hidden neurons and tanh output neurons (to bound actions) representing an action vector whose dimension depends on the task.

We train the networks with proportional Prioritized Experience Replay (PER) [2] using the Adam optimizer [30] and a learning rate of $10^{-3}$ for both the autoencoder and critic and $10^{-4}$ for the actor. We use a replay buffer of size $10^5$ and a minibatch size of 64 sampled using PER. The PER hyperparameters $\alpha$ and $\beta_0$ were set to 0.6 and 0.4 respectively. The target value network's update factor $\tau$ is set to $10^{-3}$. The reward discount $\gamma$ is 0.99. We set the intrinsic reward decay constant D to 0.1. The intrinsic reward is normalized so that it remains in the interval [0, 1]. The ITM model has the threshold $e_{max}$ as its only hyperparameter, which we set to 6.0. Five nodes, i.e. predictive models, are generated on average. All predictive models used are 2-layer fully-connected MLPs of 20 tanh hidden and 16 linear output neurons trained online with Adam optimizer. Exploratory actions are Gaussian distributed with a standard deviation of 20 degrees and a mean at the current actor's output.

The above values were determined empirically based on preliminary experiments and the following findings were obtained. Different numbers for the dense layer neurons of the actor and critic networks made no significant difference to the results. For the centre-most hidden layer of the autoencoder, we tested the performance for 8, 16, 32, and 64 neurons. By reducing from 16, as finally used, to 8, the average reward decreased to below 2.5. Increasing from 16 to 32 and 64 did not significantly change the average reward. Different learning rates were evaluated and found to slightly affect the learning performance. However, learning rates below $10^{-3}$ for training the autoencoder and critic caused slow learning convergence. Minibatch sizes larger than 64 did not lead to a considerable performance improvement. The value of $\gamma$ did not correlate with the performance.

Our own DDPG implementation for learning from pixels uses the same neural architecture described in [8] and the best-performing hyperparameters we empirically found, in addition to training with proportional PER. A comparison between the number of learnt parameters used in the proposed neural architecture (see Figure 1) and that of DDPG is presented in Table 1.

**Parameter choice analysis:**

While the structural and learning parameters of our proposed deep neural architecture is based on standard deep learning models and so their choice can be directly understood, some other parameters are less straightforward. Here, we particularly explain the role and choice of the PER, ITM and intrinsic reward decay parameters as follows:

– In PER [2], transitions are sampled from a replay buffer with probability proportional to their priorities $P(i) = \frac{p_i^\alpha}{\sum_k p_k^\alpha}$ where $p_i$ is the priority of transition $i$ represented by the absolute value of its TD-error and the

**Algorithm 1** Deep ICAC algorithm

Initialize the parameters $\{\omega, \tilde{\omega}, \theta^V, \theta^{AC}, \omega', \theta^{V'}, \tau\}$
Initialize a growing self-organizing map $M$
Initialize replay buffer $R$
**for** $e = 1$ *to #episodes do*
  Get initial state $s_1$
  **for** $t = 1$ *to #steps do*
    Select action $a_t$ from a Gaussian distribution centered at the actor's output $AC\left(\phi_{s_t} | \theta^{AC}\right)$
    Execute $a_t$ and observe $r_t^{ext}$ and $s_{t+1}$
    Update $M$ and the predictive model of the region covering $\phi_{s_t}$ using $\left(\phi_{s_t}, a_t, \phi_{s_{t+1}}\right)$
    Compute the intrinsic reward $r_t^{int}$ using Eq. 10
    Compute the total reward $r_t$ using Eq. 11
    Store $(s_t, a_t, r_t, s_{t+1})$ in $R$
    Sample a minibatch from $R$
    Perform a minibatch SGD on the loss $L^c(\omega, \tilde{\omega}) + L^v\left(\omega, \theta^V\right)$ $w.r.t.\ \omega, \tilde{\omega}, and\ \theta^V$
    Fix $\{\omega, \tilde{\omega}, \theta^V\}$ and perform a minibatch SGD on the loss $L^a\left(\omega, \theta^{AC}\right)$ $w.r.t.\ \theta^{AC}$ from only samples with positive TD-error
    Update target network parameters $\theta^{V'} \leftarrow \tau\,\theta^V + (1-\tau)\,\theta^{V'}$, $\omega' \leftarrow \tau\omega + (1-\tau)\,\omega'$
  **end for**
**end for**

**Table 1:** Comparison between the number of learning parameters of the different deep architectures used in the experiments.

| | DDPG | Deep CACLA/ICAC |
|---|---|---|
| Actor network | 36,077,399 | 403 |
| Critic network | 36,077,585 | 935,716 |
| Total | 72,154,984 | 936,119 |

exponent $\alpha$ determines the amount of prioritization used, with $\alpha = 0$ corresponding to the uniform random sampling. The larger the value of $\alpha$ the stronger is the prioritization. The prioritization introduces a bias by changing the distribution of the transitions used for learning. To compensate for the bias, importance-sampling weights are used $w_i^{PER} = \frac{1}{(N \cdot P(i))^\beta}$, where $N$ is the buffer size. Full compensation corresponds to $\beta = 1$. These weights are multiplied by the TD-error when updating the value function parameters. The bias is less significant prior to convergence, since the policy and state distribution are non-stationary. Therefore, $\beta$ is usually annealed from some initial value $\beta_0$ to reach 1 at the end of learning. We empirically found $\alpha = 0.6$ and $\beta_0 = 0.4$ to yield stable results in all our experiments.

– In ITM [29], a new node is created when the stimulus is more than a given threshold $e_{max}$ away from the nearest node. This means that $e_{max}$ determines the desired mapping resolution as it controls the growth of the ITM map. The choice of $e_{max}$ can influence the derived intrinsic reward by affecting the number of local predictive models generated. The results of setting $e_{max}$ to 6.0 were on average better than other values we experimented with. Smaller values increased the computation time without significant performance gain.
– In the combined reward signal derived in Eq. 11, the parameter $D$ controls the decay rate of the weight of the intrinsic reward component $r_t^{int}$. Reasonably small values for $D$ keep the agent more intrinsically motivated during the early stages of learning while allowing it to become gradually less intrinsically motivated as it learns more about the world dynamics and its action values. We found $D = 0.1$ as the best performing value in our experiments.

## 3.2 Environments

Here we show the experiments conducted on three robotic environments with increasing task complexity and computational demand and present the obtained results.

### 3.2.1 Vision-based learning-to-reach

We evaluate our approach on the learning-to-reach task using the V-REP robot simulator [31]. The 3-D robotic environment used in the conducted experiments is shown in Figure 4. The environment consists of a 3-DoF robot arm with a gripper attached and a red cylindrical target object. A vision sensor is used and positioned vertically above the scene to capture real-time 84 × 84 pixel RGB images of the states of the environment. Each joint of the robot can move in the angular range of $[-\frac{\pi}{2}, \frac{\pi}{2}]$. A reaching attempt is considered successful when the gripper center is within a predetermined radius from the center of the target object (the resulting target zone area is equivalent to 9% of the total reachable area of the scene). The reward function used is as follows:

$$r_t^{ext} = \begin{cases} +10 & \textit{if successful} \\ -\|c^t - c^g\| & \textit{otherwise} \end{cases}$$

where $\|c^t - c^g\|$ is the Euclidian distance between the center of the target object $c^t$ and the gripper center $c^g$. In the experiment with sparse rewards, a reward of 0 is given for unsuccessful actions.

We ran Deep ICAC, Deep CACLA, and DDPG on dense-reward and sparse-reward environments for 10K episodes with a maximum of 10 steps per episode, with the position of the target object varying randomly every episode. Training was done at the end of each episode by sampling from the replay buffer with PER and performing a minibatch SGD using Adam. For evaluating the learned policy, training was paused after every 250 episodes and a test trial was performed that includes running the policy without exploration for 20 episodes each with a different target position not included in the training. The average total (extrinsic) reward over the 20 test episodes was then reported for every test trial. We ran all the experiments on a single Nvidia GTX 1050 GPU with an average runtime of five hours per run for each of the algorithms considered.

Figure 5 shows the results of applying the algorithms to the environment in both the dense-reward and the sparse-reward settings. The results shown are averages over 20 seeds.

The performance of the learned policy was almost identical among the three algorithms during the first five test trials (1K learning episodes) in the dense-reward setting and the first ten trials (2.5K learning episodes) in the challenging sparse-reward setting. However, only the policies learned with Deep ICAC and Deep CACLA continued to improve steadily with Deep ICAC converging faster to an average return of 7.1 in the dense-reward setting and over 8.0 (*i.e.* success rate of >80%) just below the optimal policy (return of 10) in the sparse-reward setting. Despite its good performance in the dense-reward setting, Deep CACLA suffered from a premature convergence to a locally optimal policy in the sparse-reward setting. DDPG, on the other hand, showed poor stability unable to reach a good policy by the end of the training process in both reward settings.

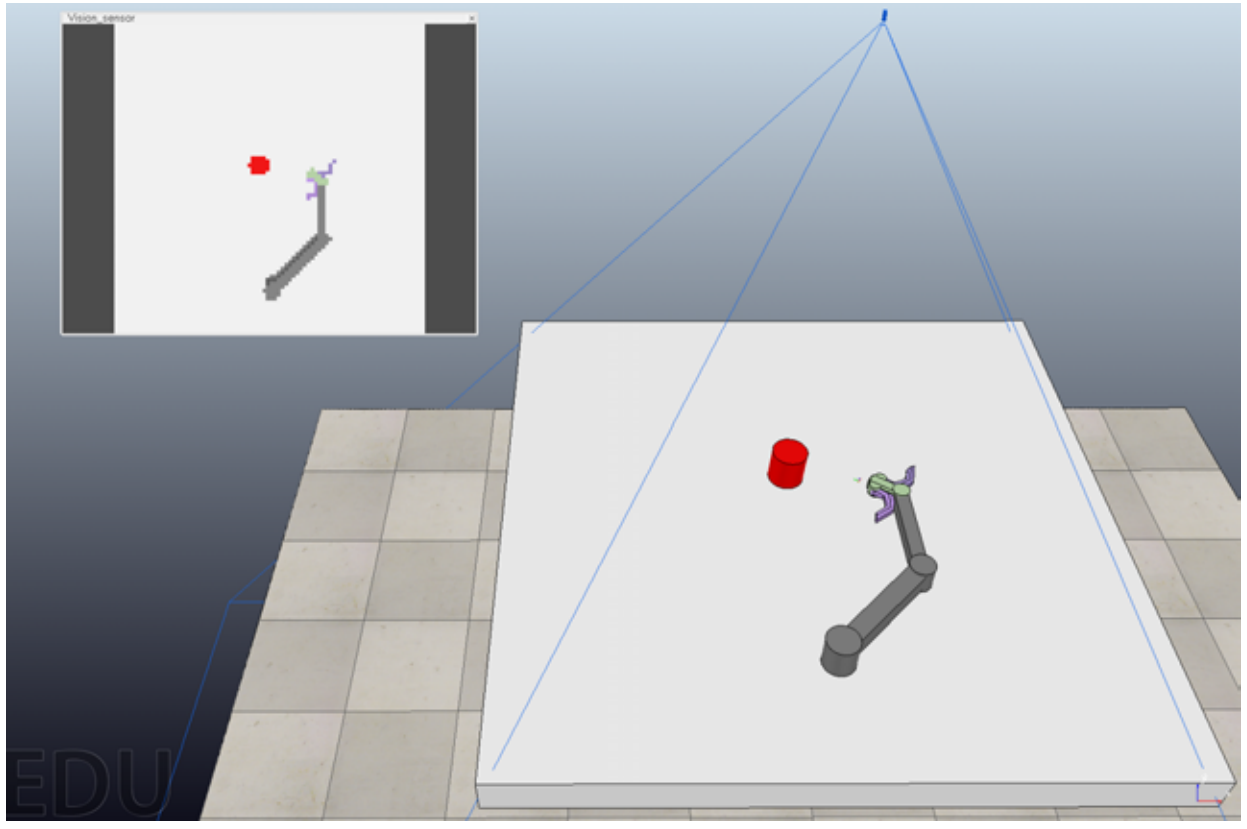

**Figure 4:** The V-REP robotic environment used in the first experiment including the 3-DoF arm with a gripper attached and a red cylindrical target. The vision sensor output is shown in the upper left corner.

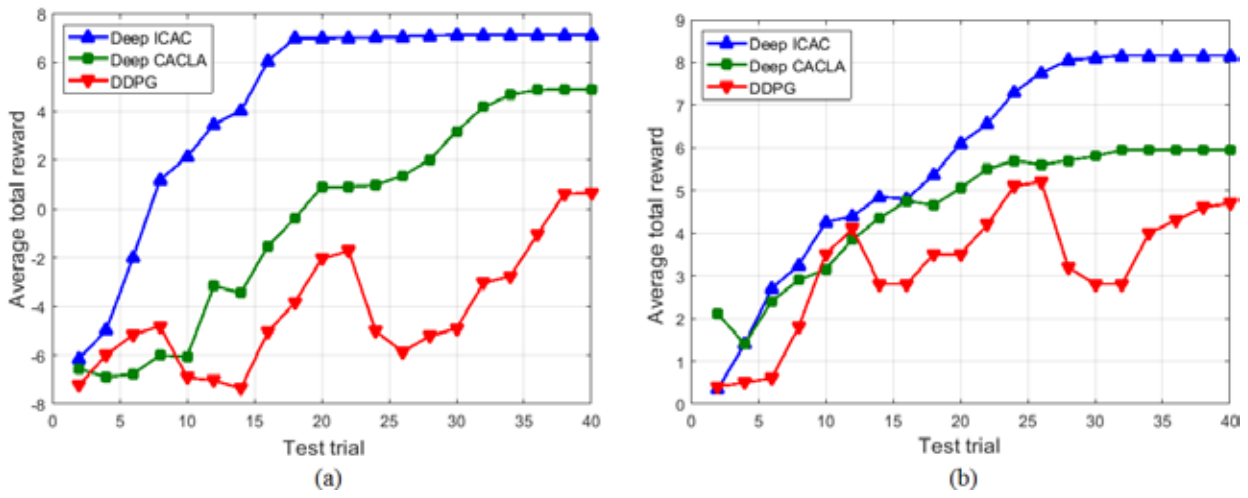

**Figure 5:** Performance curves of Deep ICAC, Deep CACLA, and DDPG on the robotic environment of the learning-to-reach task in different reward settings: dense-reward (a) and sparse-reward (b).

We also report learning statistics in terms of the average reward per episode over the entire training process (learning speed) and over the last 100 episodes of training (final performance) in Table 2. The data shown are the average over 20 runs.

### 3.2.2 Vision-based learning-to-grasp

In the second experiment, we consider robotic grasping as a learning task. Unlike reaching, grasping requires more precise motor actions, handling of external collisions with the object to grasp, and finding correct finger placement. The robotic environment consists of our Neuro-Inspired COmpanion (NICO) humanoid [32] facing a table on top of which a target object is placed. Figure 6 shows the V-REP simulation scene of the experiment.

To avoid self-collisions while allowing for a larger task space for grasp learning, we consider a motor policy involving the right shoulder joint and the right hand joints, as shown in Figure 7(a). NICO's right arm has 6 DoF of which we control one in the shoulder. The shoulder joint can move in the angular range [−100, 100] (in degrees). NICO's hand is an 11-DoF multi-fingered hand with two index fingers and a thumb each of which can move in the angular range [−160, 160] (in degrees). The robot learns to control 2 DoF: 1 DoF (shoulder joint) and 1 DoF (hand open/close). The only input to the learning algorithm is the raw data of 32×64 pixel RGB image, which is used as the state of the environment, obtained from the vision sensor output shown in Figure 7(b).

We use the following reward function:

$$r_t^{ext} = \begin{cases} +10 & \textit{if successful} \\ -10 & \textit{if object is toppled} \\ -\left\| c^t - c^h \right\| & \textit{otherwise} \end{cases}$$

where $c^t$ is the center of the target object and $c^h$ is the center of the robot hand. We determine successful grasps by moving the shoulder joint 20 degrees in the opposite direction of the recently applied joint value and measuring the Euclidean distance $\left\| c^t - c^h \right\|$ afterwards. If the distance remains below a grasp threshold of 0.04 m, the grasp is deemed successful. Otherwise, the hand is opened, the shoulder joint moves back to its previous value, and the robot continues the learning episode. In the sparse reward setting, we use the following sparse reward function:

$$r_t^{ext} = \begin{cases} +10 & \textit{if successful} \\ -10 & \textit{if object is toppled} \\ 0 & \textit{otherwise} \end{cases}$$

We run the algorithms for 10K episodes with a maximum of 50 actions per episode and with the target object randomly placed in a graspable position after every episode. The episode terminates when the object is successfully grasped, the object is toppled, or a maximum number of 20 action steps is reached.

The learning-to-grasp experiments were run on a single Nvidia GTX 1050 GPU with an average runtime of ∼25 hours per run for all the algorithms in the dense-reward setting. In the sparse-reward setting, the average runtime was 27.2, 33.8, and 35.5 hours for Deep ICAC, Deep CACLA, and DDPG respectively. Figure 8 shows the average total extrinsic reward per learning episode over five seeds.

Gradual performance improvement was observed for all the algorithms in the environment with dense reward setting, as shown in Figure 8(a). Starting at around an average total reward of −17, Deep CACLA and Deep ICAC reached a policy with an average return of 0 and 5 respectively. The DDPG progress, on the other hand, was very slow moving from −18 to −15 by the end of the learning process. In the sparse-reward environment, the algorithms were unable to make a notable progress for 3K episodes after which the learned policy of only Deep ICAC and Deep CACLA improved while DDPG's remained the same.

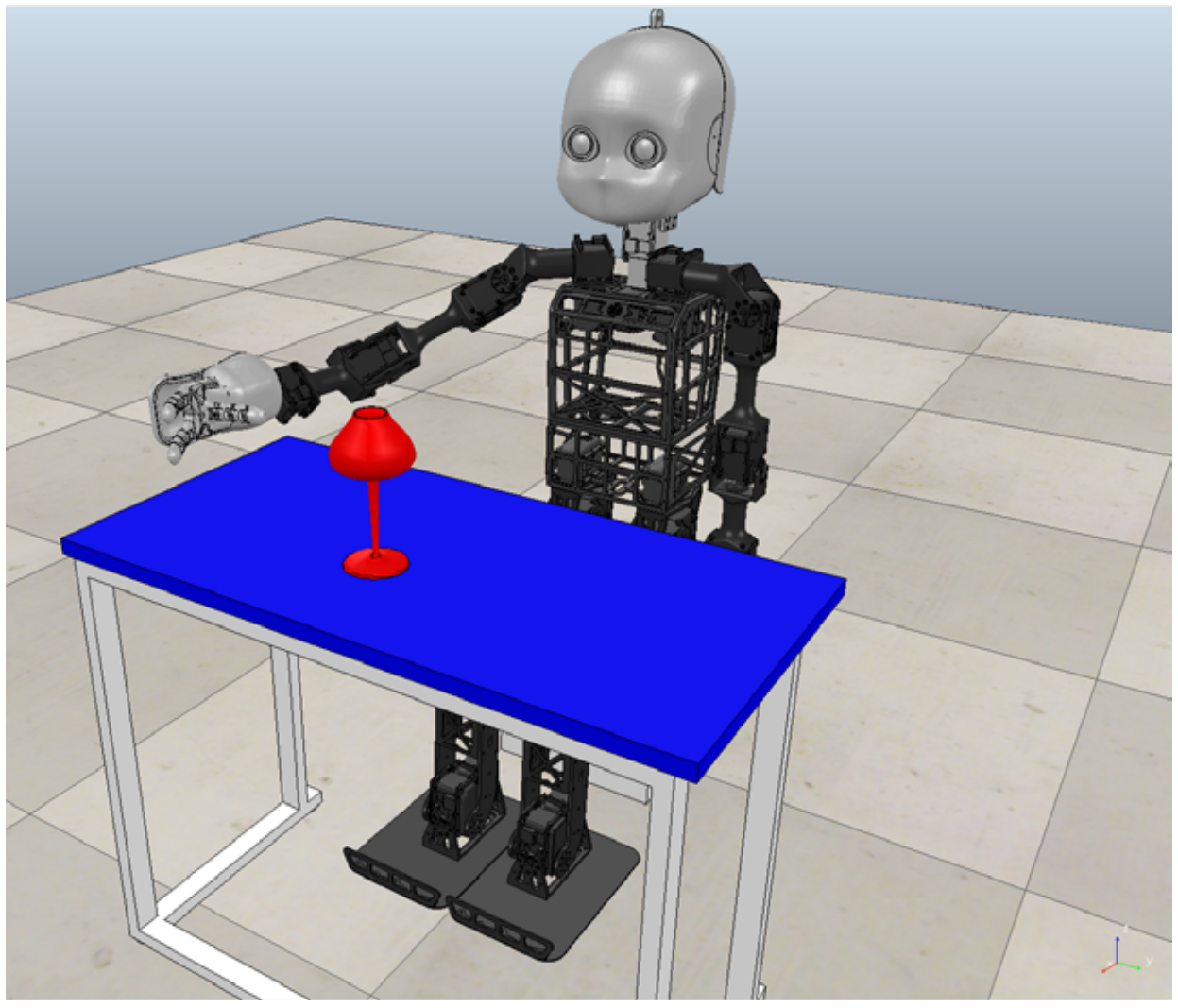

**Figure 6:** The V-REP simulation environment used in the second experiment including the NICO humanoid sitting in front of a table on top of which a target object is placed. NICO learns to grasp the object with its right multi-fingered hand.

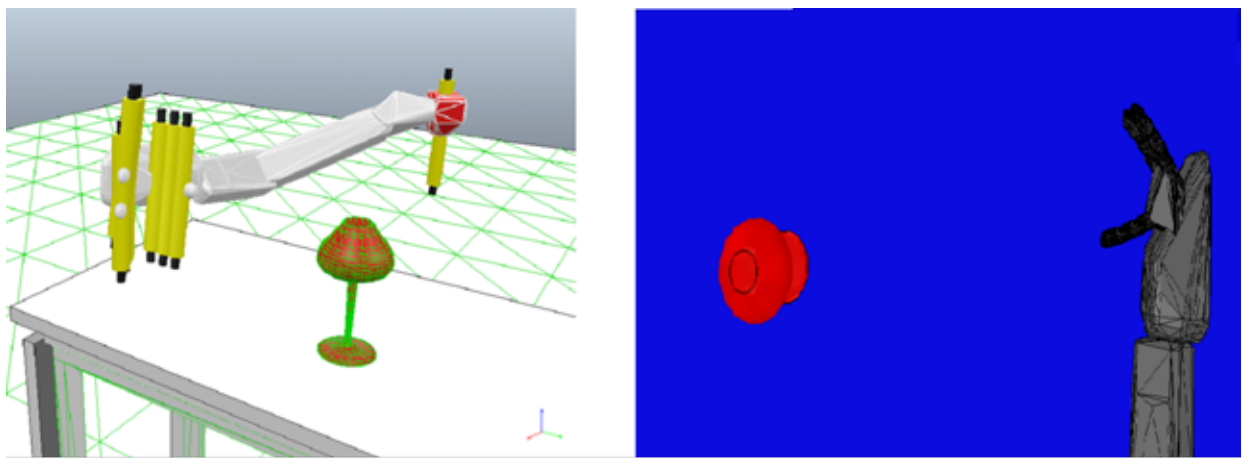

(a) (b)

**Figure 7:** The raw motor output (a) and raw sensory input (b) considered in the learning-to-grasp experiment. Yellow cylinders in (a) refer to the axes of rotations of the joints controlled during grasp learning.

### 3.2.3 Vision-based learning-to-grasp on NICO

Deep RL is well suited for research on physical, developmental robots [33]. Enabling robots to learn increasingly complex sensorimotor abilities through interaction with the real environment would move the state of the art in robotics from laborious programming tasks that can only be realized by highly specialized experts into the realm of intuitive, human-like teaching scenarios, or even robots, that can carry out repetitive learning tasks autonomously.

**Table 2:** Learning statistics in the experiments with dense rewards (upper half) and sparse rewards (lower half).

| | DDPG | Deep CACLA | Deep ICAC |
|---|---|---|---|
| Learning speed | 4.52 | **6.52** | 6.11 |
| Final performance | 4.08 | 7.01 | **7.51** |
| Learning speed | 5.34 | 7.25 | **8.14** |
| Final performance | 5.1 | 6.8 | **9.0** |

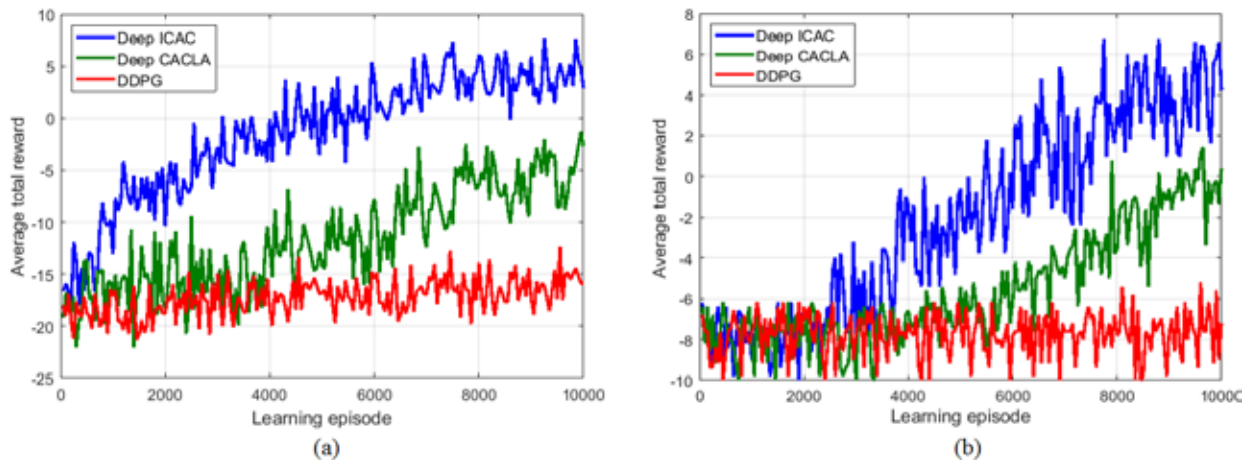

**Figure 8:** Learning curves of Deep ICAC, Deep CACLA, and DDPG on the robotic environment of the learning-to-grasp task in different reward settings: dense-reward (a) and sparse-reward (b). The average over 50 episodes is shown for readability.

To realize this, several obstacles have to be overcome: Deep RL requires a large number of samples. Successful application of deep RL has been achieved for games [1] and purely virtual environments [8]. In virtual environments, a large number of samples can be collected within a short time, without the danger of damaging the learner or to the environment and without human assistance or supervision. A simulation can be reset to its initial state, whenever an unwanted state occurs. Likewise, any required change to the environment or assistance can be automated. An example could be lifting up a toppled object and putting it back into the robot's reach. For a developing child, these chores are usually realized by its caretakers: in a typical parent-child interaction, the child learns under the supervision of adults that provide a safe environment that enables suitable learning steps.

Therefore, when moving to a real robot the research question is twofold: The core research question is the evaluation of the Deep ICAC algorithm on a real robotic system. We analyse how real sensor and motor noise affect the learning outcome. The secondary research question is the design of an experimental setup that enables autonomous learning, *i.e.* learning without constant human assistance.

As a robotic platform, we use NICO [32], a child-sized humanoid developed by the Knowledge Technology group for research on neurobotic and cognitive learning models and on human-robot interaction. NICO is an open and highly customizable platform. NICO's relevant functionalities for the experimental setup are its 6-DoF arms based on humanoid anatomy and range of motion. NICO has three-fingered HR4D hands from Seed Robotics[1] that are robust and reliable. NICO's arm is articulated with Dynamixel servomotors and controlled via the PyPot framework[2] and open NICO API by the Knowledge Technology group[3].

As the presented experiments only use the upper body functionality, the experiments are carried out on the torso version of NICO that is placed in a fixed position as if seated at a table, as shown in Figure 9. Though NICO has two integrated cameras in its head and can view its workspace on the table with its articulated head, an external camera was used to mimic the position of the virtual camera from the experiment presented in Section 3.2.2 to ensure comparability and transferability. We also successfully tested transfer of a network that has been trained on the simulator to the real NICO, but we did not use this network in the presented results. Dedicated study and analysis of the transferability of the approach is a promising area of future work.

Our physical experimental setup follows the approach by Kerzel and Wermter [34] in which a robot is able to manipulate its environment with simple, non-learned motor actions to provide suitable learning input. To learn to grasp, the robot executes a self-learning cycle depicted in Figure 9. Initially, NICO moves the hand to its start position and the grasp-learning object is put into NICO's hand (a), NICO then grasps the object and moves it to a random position on the table by using only its shoulder joint (b). The joint position is recorded and the object is released, the now empty hand moves back to the home position (c). So far, we have utilized basic robotic motor abilities, now the learning phase begins: The top-mounted camera provides an image to the learning algorithm (see Figure 10), and the

1 http://www.seedrobotics.com
2 http://github.com/poppy-project/pypot
3 http://www.inf.uni-hamburg.de/en/inst/ab/wtm/research/neurobotics/nico.html

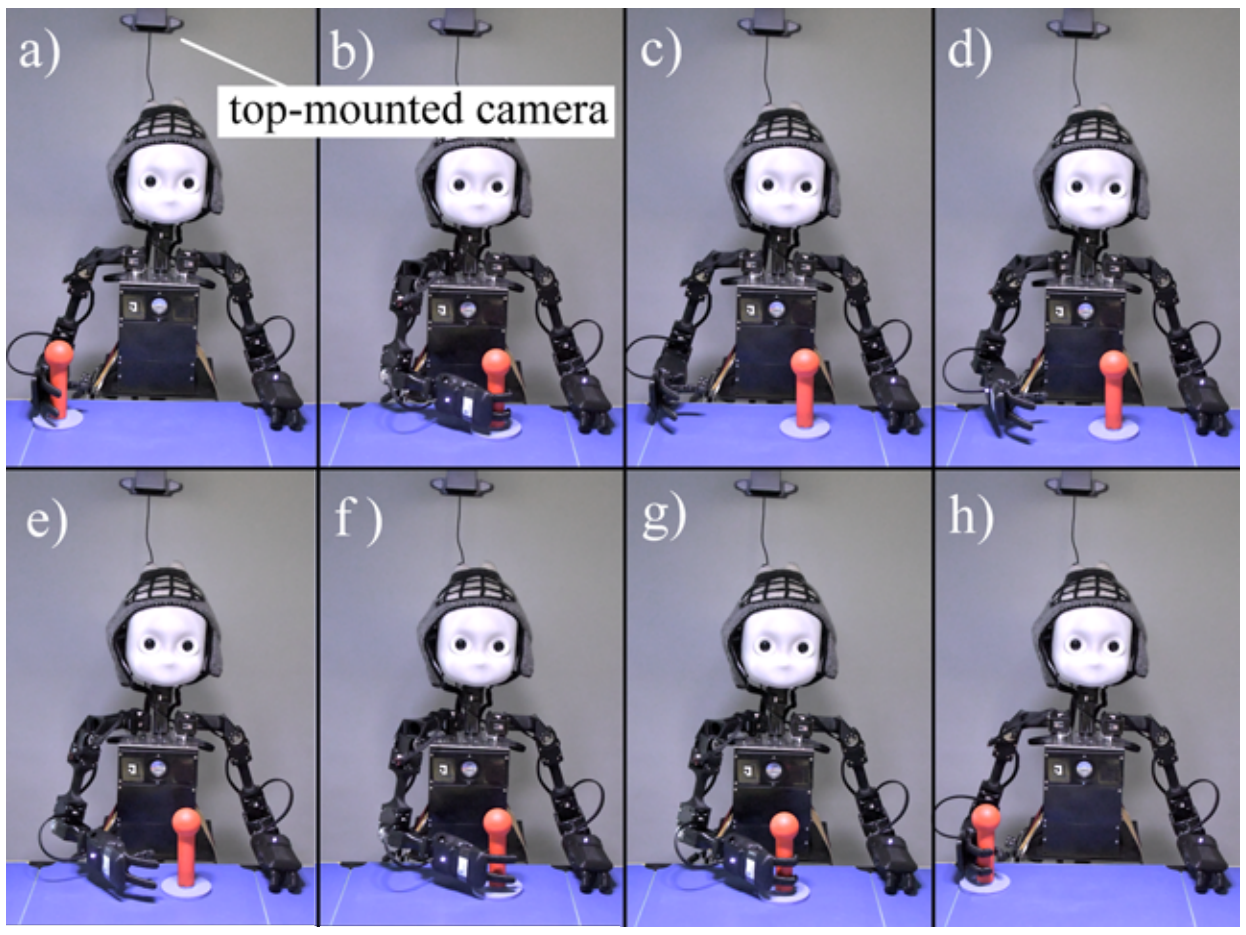

**Figure 9:** NICO experimental setup during learning including a red object for grasp-learning and a top-mounted camera. The experiment starts with NICO's hand at its start position (a). Using its shoulder joint, NICO grasps and moves the object to a random target position which is then recorded (b). Next, NICO moves back the hand to the home position (c). Learning starts by taking the image provided by the top-mounted camera as an input and producing an action output from the actor network of the Deep RL algorithm. A sequence of actions is mostly required to reach and grasp the object since the maximum angle change of the joint is limited (d-f). NICO closes its hand when the algorithm recognizes that the object has been reached (g). Once the object is grasped, the hand with the object grasped is moved to the home position and the learning cycle is repeated (h). In case of reaching a maximum of 50 action steps, the shoulder joint position is set to the recorded target position to grasp the object and move it to the home position before repeating the learning cycle.

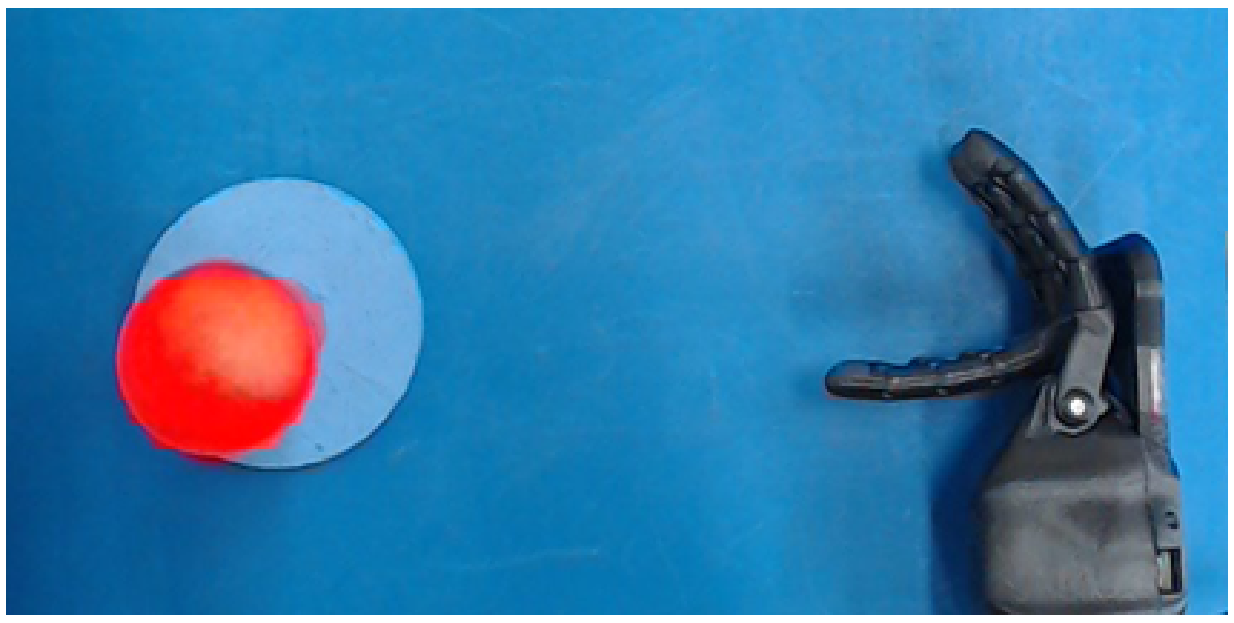

**Figure 10:** The image obtained from the top-mounted camera in the NICO experimental setup.

output of the actor's neural network is set as the next angular change of the shoulder joint. As a result, NICO moves its hand towards the grasp-learning object (d-f). As the maximum change in the joint angle is limited, mostly several steps are needed until NICO's hand reaches the object. Once the deep RL algorithm recognizes that the hand has reached the grasp-learning object based on the distance between the current and target positions of the shoulder joint, a command to close the hand is generated (g). In the case of a successful grasp, the hand and the held object are moved back to the home position (h) and the learning cycle is repeated. If a maximum number of 50 steps is reached, the hand is opened and the shoulder is moved to the recorded joint position to grasp the object which is then moved to the home position (a). We limit the joints' speed so that we do not have cases where the object is pushed away from NICO's hand or toppled over. In case the object is pushed, it stays inside NICO's open hand which is then closed on the object, once the motion is finished, and moved to the home position (a).

The advantage of this self-learning cycle is the complete independence of external assistance. Basic robotic motion and recording abilities are used to provide learning instances by placing the object at a random position as well as resetting the experiment in the cases where the learned grasp is not successful.

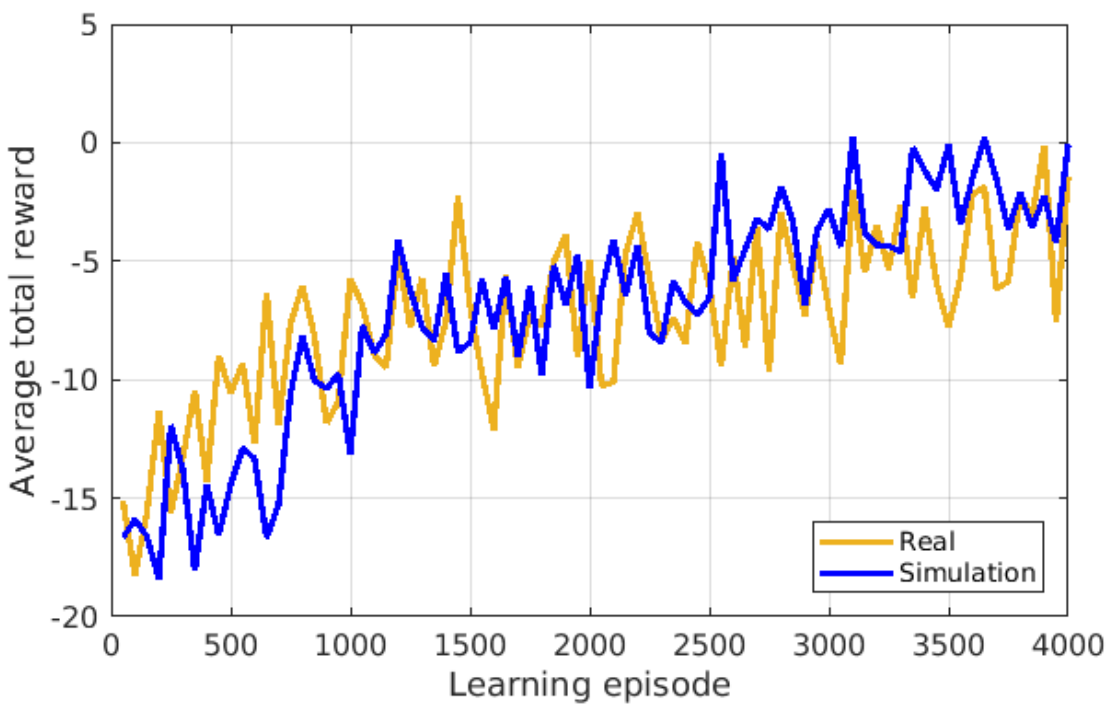

**Figure 11:** Learning curves of Deep ICAC for the vision-based leaning-to-grasp task on the simulated and real NICO humanoid.

With regard to the learning algorithm, the experiment uses the same parameters as in the virtual environment: The algorithm was trained for 4K episodes with a maximum of 50 actions per episode. A full training of the deep RL approach was conducted without human supervision for over 50 hours, during which about 15K samples were collected. During the self-learning cycle, the grasp-learning object is placed in a random graspable position within the same range of possible positions. $32 \times 64$ pixel RGB images from a top-mounted camera are used as visual input. We use the following reward function:

$$r_t^{ext} = \begin{cases} +10 & \textit{if successful} \\ -10 & \textit{if object is pushed} \\ -\left\|p^t - p^c\right\| & \textit{otherwise} \end{cases}$$

where $p^t$ and $p^c$ are the target and current positions of the shoulder joint respectively. We define a successful grasp as having a distance in the joint space of less than 1.7 degrees. All hyperparameters for the learning algorithms remain unchanged from the experiment presented in Section 3.2.2.

The results of the learning are presented in Figure 11. Compared to the training in a virtual environment both approaches show a very similar learning curve. After 4K learning episodes, the Deep ICAC on the real NICO is able to reliably grasp objects with 76% grasp accuracy (see Table 3).

**Time complexity.** One main computational difference between DDPG and our proposed algorithms is the cost of the minibatch gradient descent step during experience replay. While all the algorithms have relatively similar cost for updating the critic network, they have significantly different cost for updating the actor network. DDPG performs a product between the $1 \times s_a$ vector $\nabla_a Q\left(s, a \mid \theta^Q\right)$ and the $s_a \times s_w$ Jacobian matrix $\nabla_{\theta^\mu} \mu\left(s \mid \theta^\mu\right)$ $n$ times, where $s_a$ is the action dimension, $s_w$ is the number of the actor network's weights and $n$ is the minibatch size (see Eq. 3). This gives a complexity of $O\left(n \cdot s_a \cdot s_w\right)$. Deep CACLA and Deep ICAC, on the other hand, backpropagate the gradients of the loss in Eq. 7 computed at the actor's output layer to preceding layers with a complexity of $O\left(n \sum_{l=1}^{L} s_l s_{l-1}\right) = O\left(n \cdot s_w\right)$, where $L$ is the number of layers, $s_l$is the layer size and the input is the feature vector $\phi_s$. Since $s_w \approx 36M$ in DDPG but $s_w \approx 400$ for our actor (see Table 1), this means our actor is updated roughly 250K times faster than in DDPG when $s_a = 3$ (even more if $s_a > 3$) benefiting from the small 2-layer architecture trained on the low-dimensional $\phi_s$. The overall cost of the minibatch update is linear in the minibatch size and in the number of networks' parameters.

It should be noted that Deep ICAC has an additional cost for updating the ITM network each time a transition is observed. This involves the matching step that scales with the number of nodes and the edge adaptation step that scales with the average number of neighboring nodes. All other operations are independent of the number of nodes. The cost of updating the predictive model of the best-matching node is $O\left(\sum_{l=1}^{L} s_l s_{l-1}\right) = 640$ per transition which is the cost of a backpropagation pass on the 2-layer network. This added complexity is minimal when the average size of the ITM network is small (5 ITM nodes in our experiments). Consequently, the data efficiency of Deep ICAC does not come at the expense of a greater computational complexity, and this is especially evident since our physical robot learns in real time (Section 3.2.3).

## 4 Discussion

The results presented in Section 3 can be summarized as follows: First, Deep CACLA is significantly more stable and learns continuous control policies with high returns faster than DDPG. Second, Deep ICAC is inherently more sample-efficient than both Deep CACLA and DDPG and its superior performance is particularly pronounced in the challenging sparse-reward setting. Third, DDPG suffers from poor sample efficiency as well as learning instability,diverging from a good target policy multiple times.

The observed difference in performance between Deep CACLA and DDPG mainly stems from the policy update mechanism and the learned state representation. While DDPG updates the policy by gradient ascent on the currently learned action-value function that is initially not well trained, Deep CACLA updates the policy towards the recent action only when an actual increase in the predicted value is observed. This conservative update results in more stable learning, preventing any significant divergence from the currently best-known policy, as shown in the obtained results. The jointly optimized state representation of Deep CACLA, which is used as an input to the actor, leads to fast learning of better control policies by providing state-discriminative and value-predictive features that are low-dimensional and more accurately recognize states with high value estimates.

It is clear from the results that both DDPG and, to a lesser extent, Deep CACLA have a slow convergence to a good policy and thus require more training samples. This is largely due to the exploration policy employed which is undirected and leads to more training time spent in parts of the sensory space that are more frequently explored than others. Deep ICAC, on the other hand, provides directed, learning progress-driven exploration through its predictive model-ensemble intrinsic reward. Its intrinsic reward prevents spending additional training time in the well-explored regions of the world and is more robust to noise and task-irrelevant stochasticity in the environment. This guarantees efficient exploration and fast convergence to near-optimal policies, which is evident in the obtained results.

In the experiments on environments with sparse rewards, the robot lacks frequent feedback signals important for improving the learned policy, rendering the task more difficult. Therefore, Deep CACLA and DDPG that relay on extrinsic rewards exhibited slower learning performance in such environments than in the environments with dense rewards. Deep CACLA overcomes this difficulty by combining the sparsely available extrinsic reward with its exploration-oriented intrinsic reward, enabling the robot to continue to learn driven by the intrinsic motivation to explore.

**Table 3:** Test results of running Deep ICAC using the networks trained on the real NICO.

| | No. of trials | No. of success | Success rate |
|---|---|---|---|
| Deep ICAC on real NICO | 25 | 19 | 0.760 |

What distinguishes the learning architecture of Deep CACLA and Deep ICAC is the use of a convolutional autoencoder, rather than a standard CNN commonly used when learning control policies from raw images. A standard CNN requires either standard deep RL with reward-based losses, which is not realistic given the sparse feedback, or supervised learning with labeled pairs of states and their optimal actions. Conversely, the convolutional autoencoder can be trained unsupervised from the available images with a rich error signal, allowing seamless integration of unsupervised and RL training objectives, as detailed in Section 2.2.

The algorithms presented here learn action policies purely end-to-end without any prior knowledge or assumptions about the geometry of the robot, its environment, or the appearance of the target object in all the conducted experiments. Also, no knowledge of the kinematics of the robot and the pose of the target object is assumed. Our intrinsic reward module is general enough to be potentially used for a variety of RL methods, including value-based methods and policy gradient methods (deterministic, e.g., DDPG [8] or stochastic, e.g., A3C [10]). In the performed experiments, we use Deep CACLA for the reasons mentioned above, particularly because it provides low dimensional state representations as an input for the forward models used in generating the intrinsic reward.

We could show that the Deep ICAC algorithm enabled a physical robot to successfully learn a visuomotor ability without human assistance during the extended self-learning phase. The learned ability is limited to a single degree of freedom, but this limitation is in line with the developmental robotics paradigm of learning increasingly complex abilities, which is also found in other areas of artificial neural learning [35]. Based on the realized ability, more complex abilities can follow as each learned ability adds to the toolbox of abilities that can be used in the next learning setups.

## 5 Conclusion

We presented Deep ICAC, a fast, sample-efficient, and stable actor-critic algorithm for learning visuomotor skills in continuous action spaces. The algorithm uses a deep critic network integrated with a convolutional autoencoder and a simpler feedforward architecture for the actor. This allows the policy to be trained with maximum efficiency while learning compact, value-predictive representations. The policy in our approach is updated only from experience samples with positive Temporal-Difference error [28], which adds stability and prevents divergence when a good policy is learned. The learning progress-based intrinsic motivation of Deep ICAC supports directed and efficient exploration necessary in sparse-reward domains. The results show state-of-the-art performance of Deep ICAC for learning-to-reach and learning-to-grasp tasks in different reward settings.

In future work, we will extend the complexity of the sensorimotor task by using a visually more complex environment by introducing different backgrounds and different grasp-learning objects or also multiple objects, as realized by Eppe et al. [36]. We will also investigate the applicability of pretraining networks in a virtual environment to further increase the sample efficiency of the presented algorithm with regard to physical robot actions. Besides, the local predictive models learned in our approach as a basis for generating an intrinsic reward offer additional information on the world dynamics that is not currently used for model-based policy learning. Integrating model-based predictions with the current model-free approach also gives an interesting direction for future work.

**Acknowledgement:** This work was supported by the German Academic Exchange Service (DAAD) funding programme (No. 57214224) with partial support from the German Research Foundation DFG under project CML (TRR 169).

# References

[1] V. Mnih, K. Kavukcuoglu, D. Silver, A. A. Rusu, J. Veness, M. G. Bellemare, et al., Human-level control through deep reinforcement learning, Nature, 2015, 518(7540), 529–533

[2] T. Schaul, J. Quan, I. Antonoglou, D. Silver, Prioritized experience replay, Proceedings of the International Conference of Learning Representations (ICLR), 2016

[3] T. D. Kulkarni, A. Saeedi, S. Gautam, S. J. Gershman, Deep successor reinforcement learning, arXiv preprint arXiv: 1606.02396, 2016

[4] l T. M. Moerland, J. Broekens, C. M. Jonker, Efficient exploration with double uncertain value networks, Deep Reinforcement Learning Symposium, Advances in Neural Information Processing Systems (NIPS), 2017

[5] S. Racanière, T. Weber, D. P. Reichert, L. Buesing, A. Guez, D. J. Rezende, et al., Imagination-augmented agents for deep reinforcement learning, Advances in Neural Information Processing Systems (NIPS), 2017, 5694–5705

[6] B. C. Stadie, S. Levine, P. Abbeel, Incentivizing exploration in reinforcement learning with deep predictive models, NIPS Workshop on Deep Reinforcement Learning, 2015

[7] M. Jaderberg, V. Mnih, W. M. Czarnecki, T. Schaul, J. Z. Leibo, D. Silver, et al., Reinforcement learning with unsupervised auxiliary tasks, Proceedings of the International Conference on Learning Representations (ICLR), 2017

[8] T. P. Lillicrap, J. J. Hunt, A. Pritzel, N. Heess, T. Erez, Y. Tassa, et al., Continuous control with deep reinforcement learning, Proceedings of the International Conference on Learning Representations (ICLR), 2016

[9] G. Kalweit, J. Boedecker, Uncertainty-driven imagination for continuous deep reinforcement learning, Proceedings of the 1st Annual Conference on Robot Learning, volume 78 of Proceedings of Machine Learning Research, Mountain View, United States, 2017, 195–206

[10] V. Mnih, A. P. Badia, M. Mirza, A. Graves, T. Lillicrap, T. Harley, et al., Asynchronous methods for deep reinforcement learning, Proceedings of the International Conference on Machine Learning (ICML), 2016, 1928–1937

[11] S. Lange, M. Riedmiller, Deep auto-encoder neural networks in reinforcement learning, Proceedings of the International Joint Conference on Neural Networks (IJCNN), 2010, 1–8

[12] S. Lange, M. Riedmiller, A. Voigtlander, Autonomous reinforcement learning on raw visual input data in a real world application, Proceedings of the International Joint Conference on Neural Networks (IJCNN), 2012, 1–8

[13] C. Finn, X. Y. Tan, Y. Duan, T. Darrell, S. Levine, P. Abbeel, Deep spatial autoencoders for visuomotor learning, Proceedings of the IEEE International Conference on Robotics and Automation (ICRA), 2016, 512–519

[14] R. Legenstein, N. Wilbert, L. Wiskott, Reinforcement learning on slow features of high-dimensional input streams, PLoS Computational Biology, 2010, 6(8), p. e1000894

[15] M. B. Hafez, M. Kerzel, C. Weber, S. Wermter, Slowness-based neural visuomotor control with an Intrinsically motivated Continuous Actor-Critic, Proceedings of the 26th European Symposium on Artificial Neural Networks, Computational Intelligence and Machine Learning (ESANN), 2018, 509–514

[16] J. Schmidhuber, Formal theory of creativity, fun, and intrinsic motivation (1990–2010), IEEE Transactions on Autonomous Mental Development, 2010, 2(3), 230–247

[17] R. Houthooft, X. Chen, Y. Duan, J. Schulman, F. D. Turck, P. Abbeel, VIME: variational information maximizing exploration, Advances in Neural Information Processing Systems (NIPS), Long Beach, CA, USA, 2016, 1109–1117

[18] S. Mohamed, D. J. Rezende, Variational information maximisation for intrinsically motivated reinforcement learning, Advances in Neural Information Processing Systems (NIPS), Montréal, Canada, 2015, 2116–2124

[19] N. Chentanez, A. G. Barto, S. Singh, Intrinsically motivated reinforcement learning, Advances in Neural Information Processing Systems (NIPS), Vancouver, British Columbia, Canada, 2005, 1281–1288

[20] D. Pathak, P. Agrawal, A. A. Efros, T. Darrell, Curiosity-driven exploration by self-supervised prediction, Proceedings of the International Conference on Machine Learning (ICML), 2017, 2778–2787

[21] P. Y. Oudeyer, F. Kaplan, V. V. Hafner, Intrinsic motivation systems for autonomous mental development, IEEE Transactions on Evolutionary Computation, 2007, 11(2), 265–286

[22] M. B. Hafez, C. Weber, S. Wermter, Curiosity-driven exploration enhances motor skills of continuous actor-critic learner, Proceedings of the 7th Joint IEEE International Conference on Development and Learning and Epigenetic Robotics (ICDL-EpiRob), Lisbon, Portugal, 2017, 39–46

[23] M. B. Hafez, C. K. Loo, Curiosity-based topological reinforcement learning, Proceedings of the 2014 IEEE International Conference on Systems, Man and Cybernetics (SMC), San Diego, CA, USA, 2014, 1979–1984

[24] T. D. Kulkarni, K. Narasimhan, A. Saeedi, J. Tenenbaum, Hierarchical deep reinforcement learning: Integrating temporal abstraction and intrinsic motivation, Advances in neural information processing systems (NIPS), 2016, 3675–3683

[25] S. Sukhbaatar, Z. Lin, I. Kostrikov, G. Synnaeve, A. Szlam, R. Fergus, Intrinsic motivation and automatic curricula via asymmetric self-play, Proceedings of the International Conference on Learning Representations (ICLR), 2018

[26] A. E. Stahl, L. Feigenson, Expectancy violations promote learning in young children, Cognition, 2017, 163, 1–14

[27] C. Kidd, S. T. Piantadosi, R. N. Aslin, The goldilocks effect: Human infants allocate attention to visual sequences that are neither too simple nor too complex, PloS one, 2012, 7(5), p. e36399

[28] H. Van Hasselt, Reinforcement learning in continuous state and action spaces, Reinforcement Learning, Springer, Berlin, Heidelberg, 2012, 207–251

[29] J. Jockusch, H. Ritter, An instantaneous topological mapping model for correlated stimuli, Proceedings of the International Joint Conference on Neural Networks (IJCNN), Washington, 1999, 529–534

[30] D. Kingma, J. Ba, Adam: A method for stochastic optimization, Proceedings of the International Conference on Learning Representations (ICLR), 2014

[31] E. Rohmer, S. P. Singh, M. Freese, V-REP: A versatile and scalable robot simulation framework, Proceeding of the IEEE/RSJ International Conference on Intelligent Robots and Systems (IROS), 2013, 1321–1326

[32] M. Kerzel, E. Strahl, S. Magg, N. Navarro-Guerrero, S. Heinrich, S. Wermter, NICO – Neuro-Inspired COmpanion: A developmen-

tal humanoid robot platform for multimodal interaction, Proceedings of the IEEE International Symposium on Robot and Human Interactive Communication (RO-MAN), 2017, 113–120
[33] A. Cangelosi, M. Schlesinger, Developmental robotics: From babies to robots, Cambridge, MA: MIT Press, 2015
[34] M. Kerzel, S. Wermter, Neural end-to-end self-learning of Visuomotor skills by environment interaction, Proceedings of the International Conference on Artificial Neural Networks (ICANN), 2017, 27–34
[35] J. L. Elman, Learning and development in neural networks: The importance of starting small, Cognition, 1993, 48(1), 71–99
[36] M. Eppe, M. Kerzel, S. Griffiths, H. G. Ng, S. Wermter, Combining deep learning for visuomotor coordination with object identification to realize a high-level interface for robot object-picking, Proceedings of the IEEE-RAS International Conference on Humanoid Robotics (Humanoids), 2017, 612–617